\documentclass[runningheads]{llncs}
\usepackage[T1]{fontenc}
\usepackage{graphicx}
\usepackage{booktabs}
\usepackage[misc]{ifsym}
\newcommand{\corr}{(\Letter)}
\usepackage{mwe}
\usepackage{multirow, multicol}
\usepackage{hyperref}
\usepackage{amsmath}
\usepackage{amsfonts}
\usepackage{amssymb}
\usepackage{mathtools}
\usepackage{xcolor}

\begin{document}

\title{Leveraging External Factors in Household-Level Electrical Consumption Forecasting using Hypernetworks}

\titlerunning{Electrical Consumption Forecasting using Hypernetworks}

\author{Fabien Bernier\orcidID{0000-0002-1887-9725} \corr \and
Maxime Cordy\orcidID{0000-0001-8312-1358} \and
Yves Le Traon\orcidID{0000-0002-1045-4861}}

\tocauthor{Fabien Bernier, Maxime Cordy, Yves Le Traon}
\toctitle{Leveraging External Factors in Household-Level Electrical Consumption Forecasting using Hypernetworks}

\authorrunning{F. Bernier et al.}

\institute{
SnT, University of Luxembourg, Luxembourg \\
\email{\{fabien.bernier,maxime.cordy,yves.letraon\}@uni.lu}
}

\maketitle

\begin{abstract}

Accurate electrical consumption forecasting is crucial for efficient energy management and resource allocation. While traditional time series forecasting relies on historical patterns and temporal dependencies, incorporating external factors --- such as weather indicators --- has shown significant potential for improving prediction accuracy in complex real-world applications. However, the inclusion of these additional features often degrades the performance of global predictive models trained on entire populations, despite improving individual household-level models. To address this challenge, we found that a hypernetwork architecture can effectively leverage external factors to enhance the accuracy of global electrical consumption forecasting models, by specifically adjusting the model weights to each consumer.

We collected a comprehensive dataset spanning two years, comprising consumption data from over 6000 luxembourgish households and corresponding external factors such as weather indicators, holidays, and major local events. By comparing various forecasting models, we demonstrate that a hypernetwork approach outperforms existing methods when associated to external factors, reducing forecasting errors and achieving the best accuracy while maintaining the benefits of a global model.

\end{abstract}

\section{Introduction}

Time series forecasting has traditionally relied on historical patterns and temporal dependencies to predict future values. However, in complex real-world applications such as electrical consumption prediction, the incorporation of external factors has proven crucial for improving forecast accuracy \cite{wang2019}. These exogenous variables provide additional context that can significantly influence consumption patterns beyond what historical data alone can reveal.

In the specific case of electrical load forecasting, numerous studies have demonstrated that consumption patterns are heavily influenced by external factors such as weather conditions, calendar effects, and socio-economic indicators \cite{hong2016}. Temperature, in particular, has been shown to have a strong relationship with electricity demand, as heating and cooling needs vary significantly with ambient temperature \cite{chen2017}. Additionally, calendar variables including holidays, weekends, and seasonal patterns have been shown to capture regular variations in consumption behavior effectively \cite{wang2019}\cite{zhang2017}. These behaviors, however, are household-specific --- e.g., a household using electric heating has a consumption more sensitive to cold temperatures than a household relying on gas. This represents a challenge to global forecasting models, which therefore have to capture specific behaviors when predicting the consumption.

In order to forecast consumption, two strategies can be distinguished:
\begin{itemize}
\item \textbf{Global model:} A unified model trained on aggregated data across the entire consumer population. This centralized approach facilitates comprehensive pattern recognition across diverse consumption behaviors, enhancing generalization capabilities while minimizing computational infrastructure requirements. Furthermore, recent architectural innovations specifically address multi-channel time series~\cite{itransformer}\cite{card}.
\item \textbf{Individual models:} A dedicated model trained for each consumer entity. These specialized models capture household-specific consumption patterns with high fidelity. While traditionally resource-intensive in terms of computation and storage, recent advances in federated learning mitigate these constraints~\cite{federated-learning}, though hardware limitations for on-device machine learning deployment remain significant.
\end{itemize}

To compare these two paradigms, we assess them on real-world data provided by an industrial partner, containing more than 6000 households consumptions over two years and corresponding external factors, ranging from weather data to football\footnote{\textit{Football} in this paper refers to \textit{``soccer''}} events.
As our results later demonstrate, although incorporating external factors as features theoretically enhances performance, these lead to overall performance degradation in global models. Conversely, individual models excel at mapping external factors to consumer-specific responses, but introduce substantial computational and storage overhead that scales linearly with the consumer population.
In particular, this approach fails to capitalize on the substantial behavioral similarities across consumers. Since many households share comparable consumption patterns~\cite{wang2016clustering}, training completely separate models results in significant parameter redundancy, as each individual model essentially learns the same forecasting task (electricity consumption) with variations to accommodate specific consumer profiles. This redundancy wastes computational resources and misses opportunities for knowledge sharing across similar consumer segments.

\begin{figure}[!ht]
    \centering
    \includegraphics[width=\linewidth]{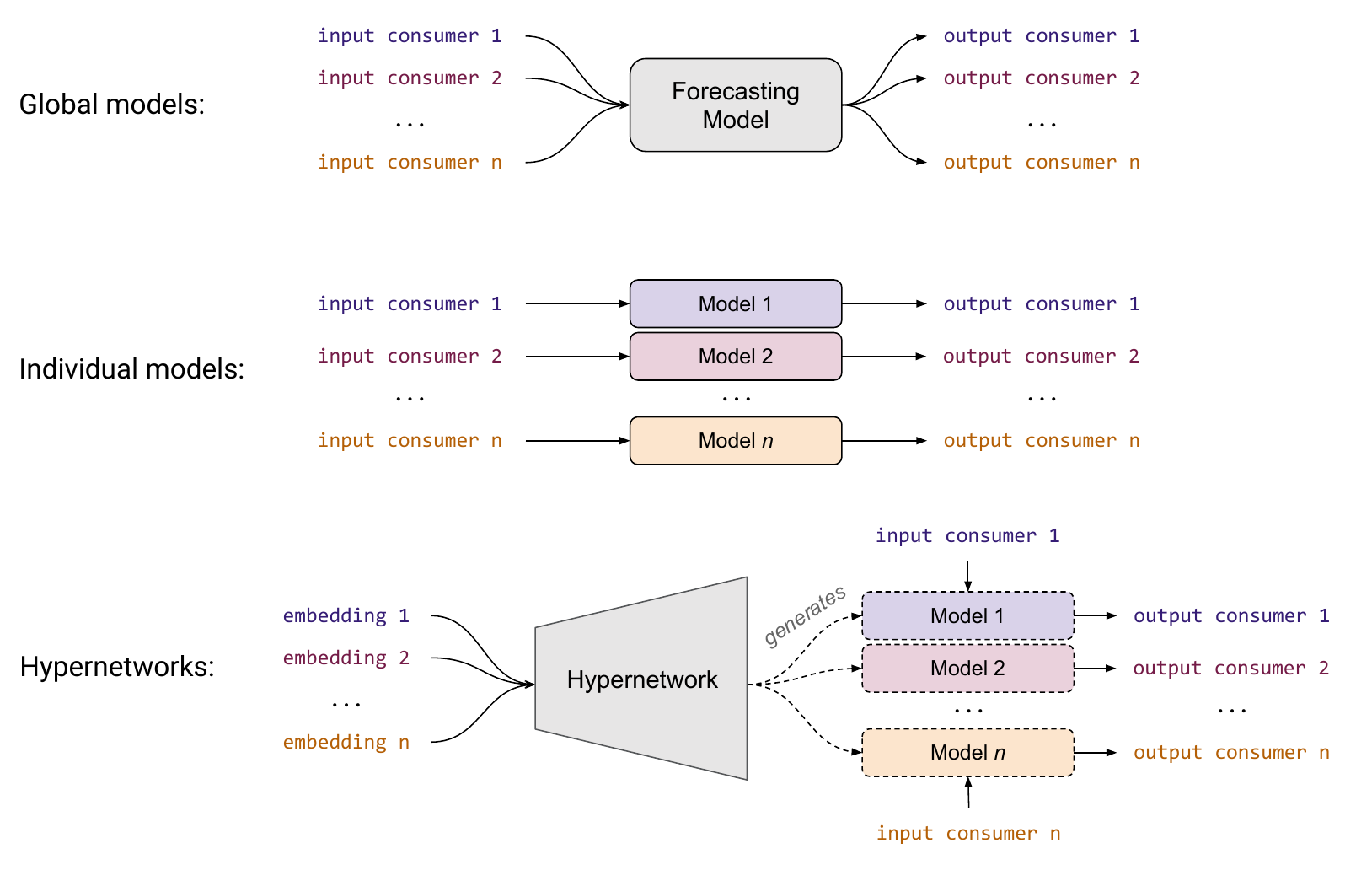}
    \caption{Difference between global and individual models, and the proposed in-between solution using hypernetworks.}
    \label{fig:paradigms}
\end{figure}

In order to bridge the gap between global models efficiency and individual models precision, hypernetworks offer a promising architectural paradigm, illustrated in Figure \ref{fig:paradigms}. Hypernetworks~\cite{hypernetworks} are meta-models designed to generate the weights of a primary task network conditioned on specific inputs. In our context, a hypernetwork can dynamically produce customized parameters for each consumer based on their unique embedding and current situation. This approach maintains the personalization advantages of individual models while dramatically reducing the parameter space, rather than maintaining thousands of separate forecasting models.

In this paper, we introduce a novel approach using hypernetworks and consumer-specific embeddings that enable global models to differentiate between individual households. These compact embeddings require minimal storage compared to full individual model parameters while preserving household-specific information. Our experimental results demonstrate that the hypernetwork architecture is the only one in the tested benchmark to leverage external factors to reduce forecasting error --- and ultimately get the lowest error, beating state-of-the-art models by up to 16\% --- while conventional approaches result in performance degradation. This improvement enables more accurate, individualized forecasting within a computationally efficient framework.

\section{Background}

Time series forecasting has evolved from classical statistical methods to advanced deep learning architectures. Traditional approaches like ARIMA~\cite{arima} rely on temporal dependencies within univariate series but struggle with incorporating exogenous variables effectively. More recently, neural network-based models have demonstrated significant improvements in handling complex time series tasks.

Transformer-based architectures~\cite{transformers} have been adapted for time series forecasting, with models like Informer~\cite{informer} addressing the quadratic complexity limitations of vanilla transformers. N-HiTS~\cite{nhits} extends the interpretable N-BEATS framework~\cite{nbeats} by introducing hierarchical interpolation and multi-rate data processing for improved performance across multiple horizons.

When it comes to electricity load forecasting, a critical challenge is to effectively incorporate multiple information channels, including historical consumption and various exogenous factors. Recent architectures specifically target this \textit{multivariate} challenge: iTransformer~\cite{itransformer} revolutionizes time series modeling by treating individual features as tokens and timestamps as channels, inverting the traditional approach. PatchTST~\cite{patchtst} applies patching strategies to decompose time series into subseries, enabling more robust feature extraction. Lately, CARD~\cite{card} introduced channel attention mechanisms that dynamically weight the importance of different input variables.

These models however still have to process the input time series to figure out the consumer's profile, which can be highly different from one time series to another. Additionally, recognizing consumers profiles may also require longer input time series (e.g. in order to analyze their behaviors during vacations). Mixture of Experts (MoE) models~\cite{moe} offer another approach to handling heterogeneous patterns in time series data. These architectures dynamically route inputs to specialized subnetworks, allowing the model to develop expertise given a specific embedding. Mixture of Linear Experts (MoLE)~\cite{mole} extends this concept by creating embeddings that represent input characteristics in order to create this embedding, further improving adaptability to diverse time series behaviors.

Hypernetworks~\cite{hypernetworks} represent a powerful paradigm where one network generates the weights for another. In the domain of time series, this approach has shown particular promise for addressing distribution shifts in time series~\cite{hypernetworks-shift} and has been applied to implicit neural representations as demonstrated in HyperTime~\cite{hypertime}. Hypernetworks are especially relevant for our work as they can efficiently generate consumer-specific parameters from compact embeddings, potentially capturing individual household behaviors without requiring separate models for each consumer.

In the context of electricity load forecasting, these architectural innovations offer promising directions for improving prediction accuracy while maintaining computational efficiency. Our work builds upon these foundations to address the specific challenges of capturing consumer-specific responses to exogenous factors.

\section{Hypernetworks for Time Series Forecasting}

\subsection{Problem Formulation}

We address the task of forecasting electrical consumption time series for a diverse set of consumers while incorporating various external factors. Let $\mathcal{X} = \{x_1, x_2, \ldots, x_N\}$ represent the set of $N$ consumer entities, each with its own hourly electrical consumption time series. For each consumer $x_i$, we denote its consumption at time $t$ as $x_{i,t} \in \mathbb{R}$. Additionally, we have a set of numerical external factors $\Phi = \{\phi_1, \phi_2, \ldots, \phi_k\}$ (additional time series, such as temperature) and categorical external factors $\mathcal{C} = \{c_1, c_2, \ldots, c_m\}$.

Our objective is to predict future consumption values $y_{i,t:t+h} \coloneqq x_{i, t+L:t+L+h}$ for a horizon $h$ for every consumer $i$, given historical consumption $x_{i,t:t+L}$ for an input length $L$ and external factors $\Phi_{t:t+L}$ and $\mathcal{C}_{t:t+L}$.

\subsection{Model Architecture}

\begin{figure}[!ht]
    \centering
    \includegraphics[width=\linewidth]{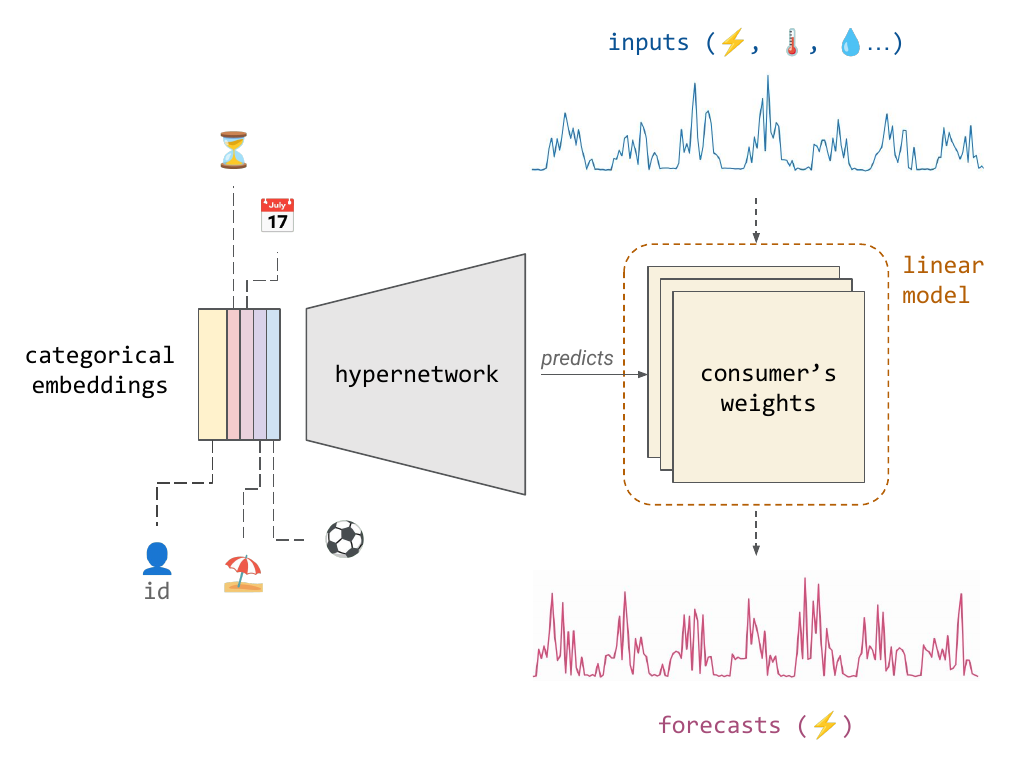}
    \caption{Overview of the hypernetwork pipeline}
    \label{fig:overview}
\end{figure}

Our proposed architecture consists of three main components: (1) an embedding layer for categorical variables, (2) a hypernetwork that generates consumer-specific weights, and (3) a linear forecasting model with these consumer-specific weights. The hypernetwork itself can be seen as a weights generator --- that outputs matrices for the linear model --- and essentially shares the same architecture than an image decoder~\cite{vqvae}. An overview of the pipeline is illustrated in Figure \ref{fig:overview}.

\subsubsection{Embedding Representation for Categorical Variables.}

For each categorical external factor $c_j \in \mathcal{C}$, we learn a dense embedding representation:

\begin{equation}
\mathbf{e}_j = \text{Embed}(c_j) \in \mathbb{R}^{d_j}
\end{equation}

where $d_j$ is the embedding dimension for factor $j$. Specifically, when categorical features are related and complementary, we sum their embeddings as follows:

\begin{equation}
    \mathbf{e}_\text{event} = 
    \begin{cases} 
        \mathbf{e}_{\text{no event}}, & \text{if} \ c_{\text{event}_k} = 0 \text{ for all } k \\
        \sum_{k \in \{k | c_{\text{event}_k} = 1 \}} \mathbf{e}_{\text{event}_k}, & \text{otherwise}
    \end{cases}
\end{equation}

\begin{figure}[!ht]
    \centering
    \includegraphics[width=1.02\linewidth]{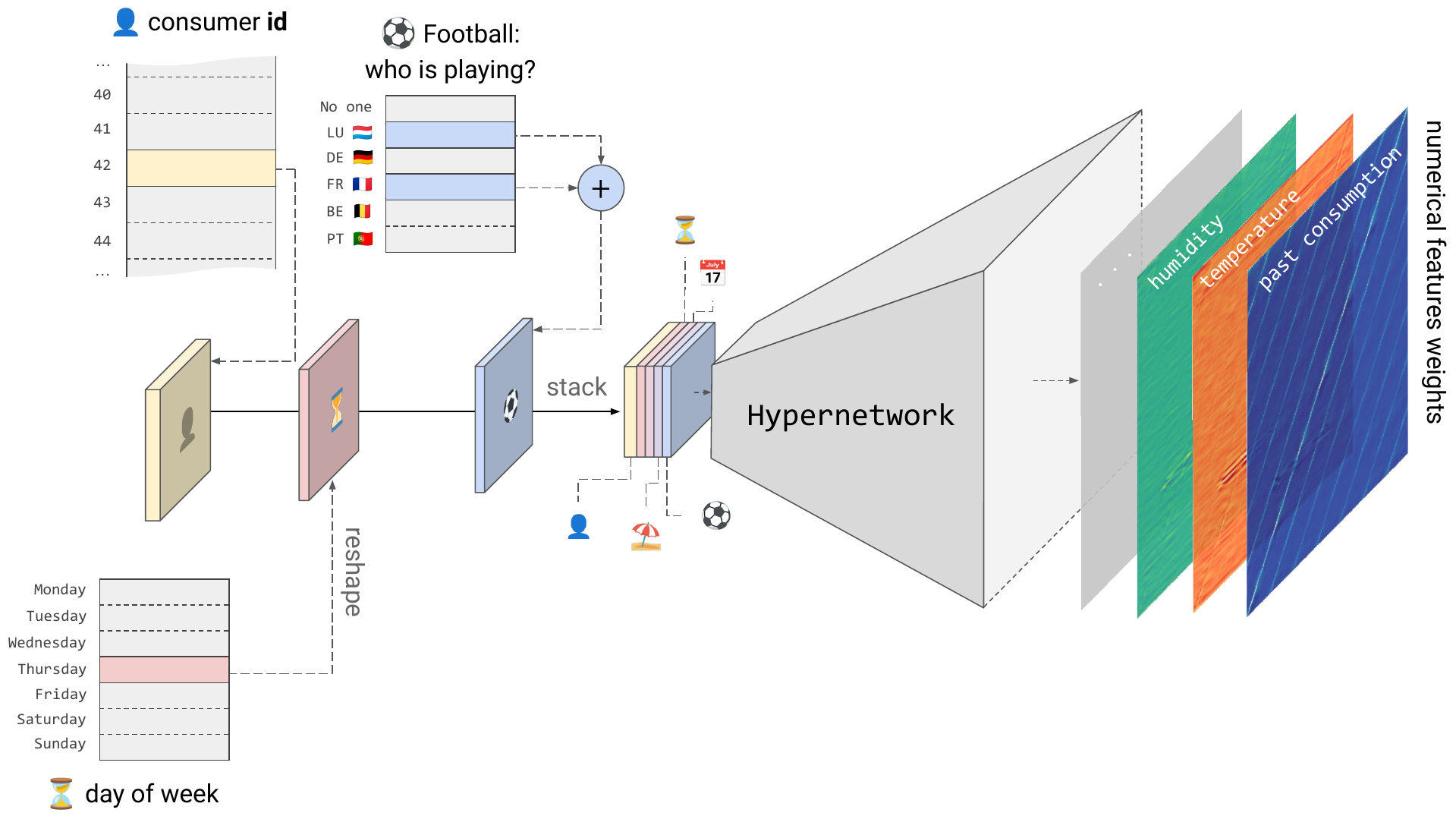}
    \caption{Illustration of example embeddings. Each consumer ID and other known categorical features are transformed to embeddings, which are reshaped and stacked together to form the Hypernetwork input.}
    \label{fig:embeddings}
\end{figure}

All categorical embeddings are reshaped to matrices of size $(p, q)$ and stacked together to form the hypernetwork input, as shown in Figure \ref{fig:embeddings}. The resulting input tensor is denoted $\mathbf{z}_{i,t}$. The output matrices predicted by the hypernetwork have proportional dimensions from the inputs, and are of shape $(p \times u, q \times u)$, where $u$ is the upscaling factor.

\subsubsection{Forecasting Mechanism.}

The hypernetwork $H_\theta$ with parameters $\theta$ takes the concatenated features $\mathbf{z}_t$ and generates the weights for a consumer-specific linear forecasting model:

\begin{equation}
\mathbf{W}_{i,t} = H_\theta(\mathbf{z}_{i,t}) \in \mathbb{R}^{L \times h \times p}
\end{equation}

where $p = k+1$ is the input dimension to the linear model, corresponding to the number of input time series; $L$ is the input length, and $h$ is the forecast horizon.

The consumer-specific weights $\mathbf{W}_{i,t}$ are then used in a linear model to produce the final forecasts. For each consumer $i$ at time $t$, the input to the linear model includes both historical consumption values $x_{i,t:t+L} \in \mathbb{R}^{L}$ and the numerical external factors $\Phi_{t:t+L} \in \mathbb{R}^{k \times L}$.

The forecast for the next $h$ time steps is then computed as:

\begin{equation}
\hat{\mathbf{y}}_{i,t:t+h} = \mathbf{W}_{i,t} \cdot \begin{pmatrix} x_{i,t:t+L} \\ \phi_{1,t:t+L} \\ \ldots \\ \phi_{k, t:t+L} \end{pmatrix}
\end{equation}

\subsubsection{Loss Function and optimization.}

We jointly optimize the hypernetwork parameters $\theta$ along with all categorical feature embeddings $e_i$ by minimizing the Mean Squared Error (MSE) between predictions and ground truth:

\begin{equation}
\min_{\theta, \{e_i\}} \sum_{i=1}^{N} \sum_{t \in \mathcal{T}} \|\hat{\mathbf{y}}_{i,t:t+h} - \mathbf{y}_{i,t:t+h}\|^2
\end{equation}

where $N$ is the number of consumers, $\mathcal{T}$ is the set of time points in the training data, $\hat{\mathbf{y}}_{i,t:t+h}$ represents the predicted values, and $\mathbf{y}_{i,t:t+h}$ represents the ground truth values.

It is worth noting that unlike traditional neural networks where weights are directly optimized, in our approach, the hypernetwork parameters $\theta$ are optimized such that they can generate effective consumer-specific weights $\mathbf{W}_{i,t}$ for the linear forecasting model. This approach allows the model to dynamically adapt to different consumers' consumption patterns while leveraging shared knowledge across the entire consumer base.

\subsection{Experimental setup}

\subsubsection{Dataset.}
We collected a dataset comprising hourly data over a period of two years (2020 and 2021):

\begin{itemize}
    \item \textbf{Numerical features:}
    \begin{itemize}
        \item $x$: Consumption data (kWh) for $N=6,010$ households and businesses in Luxembourg, provided by the national grid operator;
        \item $\phi_\text{temp}, \phi_\text{hum}, \phi_\text{wind}, \phi_\text{sun}$: Weather indicators --- temperature (°C), humidity (\%), wind speed (km/h), sunlight (minutes of sun within one hour)\footnote{For simplicity purposes, these indicators are global for all consumers (weather in Luxembourg City), as the geographical area of study is small with few local variations.};
    \end{itemize}
    \item \textbf{Categorical features:}
    \begin{itemize}
        \item $i$: Consumer ID, ranging from 0 to 6,009;
        \item $c_\text{hour}, c_\text{dw}, c_\text{dm}, c_\text{month}$: Timestamps data --- hour of day (24 values), day of week (7), day of month (31), month of year (12);
        \item $c_\text{sh}, c_\text{ph}$: School holiday indicator (boolean), public holiday (boolean);
        \item $c_{\text{team}_1}, \ldots, c_{\text{team}_5}$: 5 booleans, indicating wether Luxembourg, Germany, France, Belgium or Portugal will be playing in the current day or not --- which are relevant teams for the studied region.
    \end{itemize}
\end{itemize}

As it is usual for electric load forecasting~\cite{gasparin2022deep}, we set a forecast horizon of 1 week ($h=168$), from an input length of 2 weeks ($L=336$). We compare results with and without the inclusion of external factors, and run further experiments where only the consumer ID is provided in addition to electrical consumption.
The dataset is partitioned chronologically into train/validation/test sets with standard 70\%/10\%/20\% ratios following established time series forecasting protocols~\cite{autoformer}. We preprocess the data by standardizing consumption values, temperature, and wind speed, while applying min-max normalization to humidity and sunlight variables as these represent naturally bounded quantities.

\subsubsection{Hyperparameters.} We set the upscaling factor $u$ to 24, which fits with the daily seasonality characteristics of electricity consumption. Given this factor and the needed sizes of the output matrices ($336 \times 168$), we have to make inputs of size $14 \times 7$. To achieve this, we concatenate two $7 \times 7$ matrices, leading to 49-dimensional vectors. One reason for this choice is the flexibility this concatenation offers: one could easily change the number of weeks in the input length or forecast horizon by getting shapes of $7a \times 7b$. For consumer IDs, we allocate twice the embedding capacity ($7 \times 7 \times 2$) to capture the more complex behavioral patterns associated with individual users. These embedding tensors are concatenated along the channel dimension before being processed by the model through four residual blocks, ultimately generating weight matrices of dimension $336 \times 168$ that map input sequences to forecast horizons. Experiments are repeated 10 times to reduce randomness effects.

\subsubsection{Baseline models.}
One natural additional solution to experiment with is Mixture of Linear Experts~\cite{mole}, as they demonstrate strong performance in general time series forecasting. Especially, each expert can specialize in specific groups of consumers, and embeddings can simply be used to attribute expert importance. We consider three MoLE variants, MoLE\_DLinear, MoLE\_RLinear and MoLE\_RMLP, the latter consisting in two dense layers expert models. 16 experts are used, as this setting allowed the good performance shown in \cite{mole}. When using categorical features, we use the same embeddings as for hypernetworks, which are then linearly mapped to a probability distribution vector that assigns experts importance.

Baseline models also include state-of-the-art forecasting models with a focus on multiple channels processing: iTransformer (2024 \cite{itransformer}), CARD (2024 \cite{card}), NHits (2023 \cite{nhits}), PatchTST (2022 \cite{patchtst}), RLinear (2022 \cite{rlinear}). For completeness, we include ARIMA as a classical statistical baseline which, despite its computational complexity, often provides competitive performance for structured time series forecasting tasks. Since the baseline models are designed for continuous multivariate time series, we adapt categorical features for fair comparison. For most categorical variables, we employ one-hot encoding to create additional binary channels. However, for the high-cardinality consumer ID feature, this approach would create an impractical number of channels. Instead, we learn low-dimensional embeddings for consumer IDs and repeat these embeddings across the temporal dimension, maintaining consistent representation while controlling dimensionality. The code is available on Github\footnote{\url{https://github.com/serval-uni-lu/hypernetworks-time-series}}.

Finally, we compare these results with individual RLinear models being trained for every individual consumer --- not predicted by the hypernetwork --- in contrast with global models cited above.

\subsubsection{Infrastructure.} We use a Quadro RTX 8000 49GB GPU for all the experiments.

\section{Results}

\begin{table}[h]
    \centering
    \setlength{\tabcolsep}{3pt}
    \caption{MSE and MAE values for different models and datasets. Models denoted with an asterisk * are not meant to handle categorical features: the consumer's ID embedding is provided in additional time series channels}
    \begin{tabular}{l c c c c c c c c}
        \toprule
        \multirow{2}{*}{{Model}} & \multicolumn{2}{c}{{No external factor}} & & \multicolumn{2}{c}{{Consumer ID only}} & & \multicolumn{2}{c}{{+ External factors}} \\ \cmidrule(lr){2-3} \cmidrule(lr){5-6} \cmidrule(lr){8-9}
        & \textbf{MSE} & \textbf{MAE} & & \textbf{MSE} & \textbf{MAE} & & \textbf{MSE} & \textbf{MAE} \\ \midrule
        Our method  &   -    &   -    & & 0.1771 & 0.1872 & & \underline{0.1734} & \underline{0.1805} \\ 
        MoLE\_DLinear & 0.1788 & 0.1899 & & 0.1798 & 0.1891 & & 0.1807 & 0.1904 \\
        MoLE\_RLinear & 0.1786 & 0.1836 & & 0.1795 & 0.1844 & & 0.1787 & 0.1839 \\
        MoLE\_RMLP    & 0.1774 & 0.1820 & & 0.1788 & 0.1832 & & 0.1778 & 0.1826 \\
        Individual RLinears*  &   -    &   -    & & 0.1741 & 0.1819 & & \textbf{0.1725} & \textbf{0.1792} \\
        RLinear*       & 0.1806 & 0.1901 & & 0.1874 & 0.1990 & & 0.1888 & 0.2044 \\
        iTransformer*  & 0.1834 & 0.1867 & & 0.1866 & 0.1894 & & 0.1969 & 0.1966 \\ 
        CARD* & 0.1759 & 0.1816 & & 0.1760 & 0.1817 & & 0.1765 & 0.1822 \\ 
        NHits* & 0.1757 & 0.1849 & & 0.1759 & 0.1851 & & 0.1763 & 0.1854 \\ 
        PatchTST* & 0.1759 & 0.1817 & & 0.1762 & 0.1822 & & 0.1768 & 0.1856 \\
        ARIMA         & 0.1780 & 0.1893 & &   -    &   -    & &   -   &  -   \\
        \bottomrule
    \end{tabular}
    \label{tab:results}
\end{table}

Our experimental results demonstrate several key findings regarding the performance of various time series forecasting models, as shown in Table \ref{tab:results}. The comparison across different input configurations yields important insights for model selection and deployment in real-world scenarios. The standard error is always < $10^{-4}$ in the table, with two minor exceptions. More detail is provided in appendix.

\subsection{Impact of External Factors}
Perhaps the most surprising finding is that incorporating external factors generally degrades model performance across almost all architectures. This contradicts the common assumption that additional information should improve predictive accuracy. Only individual models and our hypernetwork approach exhibit improved performance when leveraging external factors, with decreases in both MSE and MAE compared to using consumer ID only or no external factors.

This exceptional behavior of hypernetworks suggests they possess a unique ability to effectively filter and use external information without introducing additional noise or complexity that harms prediction accuracy. The architecture's approach to handling multiple input channels appears fundamentally more effective than competing methods.

\paragraph{Consumer ID Embeddings.}
The performance when using only consumer ID embeddings as additional channels provides insights into how different models handle the introduction of this information. The MoLE models are the only global ones to improve the forecasting quality with the consumer ID provided --- they are, however, with the hypernetworks, the only models designed to handle this specific input.
Models not explicitly designed for this purpose always show a small degradation of performance. Despite not being optimized for categorical features, Transformer models still perform reasonably well in this scenario.

\subsection{Performance Across Model Architectures.}
The Hypernetwork architecture exhibits superior performance compared to other models by successfully \textit{imitating} the individual models approach and getting closer to its final performance, achieving the second lowest MSE (0.1734) and MAE (0.1805) when incorporating external factors. This represents a notable improvement over traditional approaches and even other deep learning models. CARD and NHits follow closely behind, with NHits demonstrating particularly strong performance (MSE: 0.1763, MAE: 0.1854), making it a viable alternative when no external factors are available.

Interestingly, the classical Arima model (MSE: 0.1780, MAE: 0.1893) remains competitive despite being significantly less complex than the deep learning approaches. This suggests that for certain forecasting tasks, traditional statistical methods should not be dismissed outright.

\subsection{Cost}

\paragraph{Training time.} Our hypernetwork approach achieves a favorable trade-off between computational resources and prediction accuracy. While generating consumer-specific weights introduces additional computational overhead during training compared to global models, this cost is substantially lower than training individual models for each consumer. Specifically, our approach reduces training time by 7 hours (approximately 70\%) compared to individualized RLinear models.

\paragraph{Memory.} The memory efficiency of our approach is particularly notable. The consumer embeddings require only 589K parameters (2.4MB), whereas individual linear models for all 6,010 consumers demand 3.392 billion parameters. This represents a parameter reduction factor of over 5,700$\times$. Extrapolating to a real-world deployment with 1 million consumers, our approach would require only megabytes of storage compared to approximately 2.3TB for individual models. This dramatic reduction in model size not only decreases storage requirements but also eliminates the significant I/O overhead that would occur when loading individual models from disk during inference --- a practical consideration not captured in our GPU-only-based timing experiments.

\begin{figure}[!ht]
    \centering
    \includegraphics[width=0.9\linewidth]{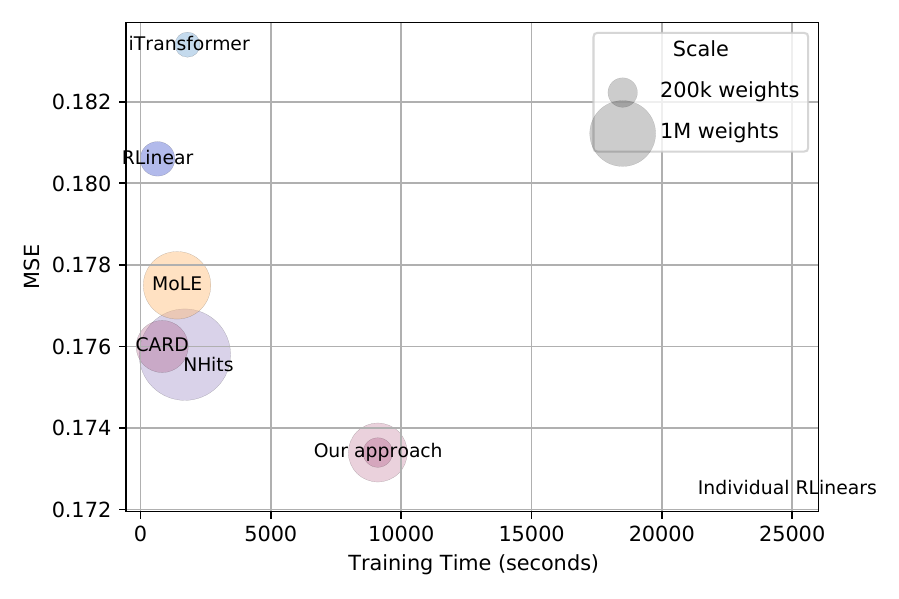}
    \caption{Comparison of models training time vs. the best resulting MSE. Bubble sizes refer to the number of weights. For our approach, we distinguish the size of the hypernetwork itself, and the size including all 6010 consumers embeddings. The size for individual RLinear models (bottom right) is not shown as it would fill in all the figure.}
    \label{fig:sizes}
\end{figure}

\subsection{Generalizing consumers embeddings}

\begin{figure}[!ht]
    \centering
    \includegraphics[width=0.7\linewidth]{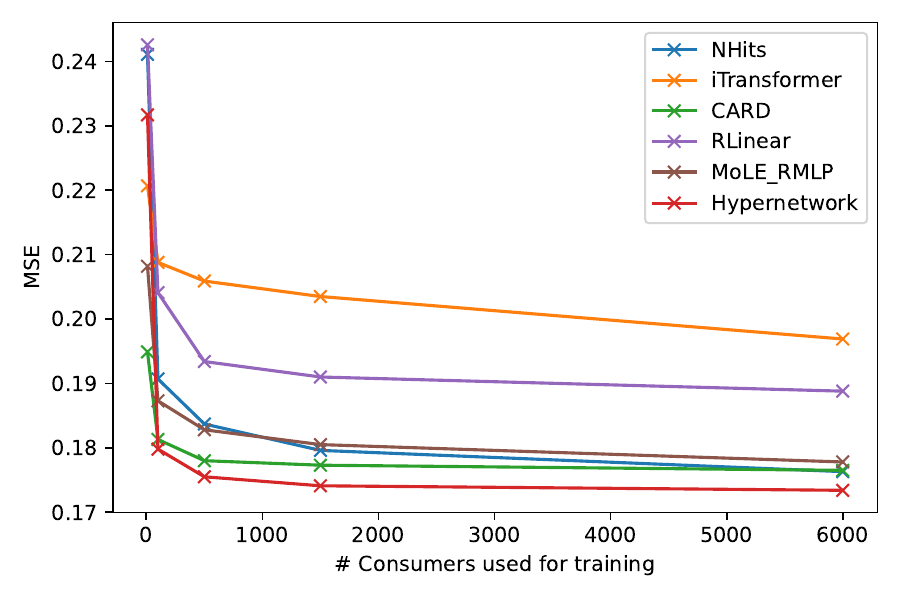}
    \caption{Performance evaluated on the full dataset when only a portion of the 6010 consumers is used to train models}
    \label{fig:performance-consumers}
\end{figure}

As consumers might evolve over time, with new ones arriving and others leaving, embeddings often need to be updated. This can be achieved by optimizing the embeddings in order to reduce the final forecasting error. One advantage of this method is that this task can be easily parallelized, and the hypernetwork model itself doesn't necessarily need to be retrained. Figure \ref{fig:performance-consumers} reveals that our hypernetwork approach, when trained on merely 8\% of the consumer base (500 out of 6010 consumers), outperforms competing models across the entire dataset, given consumers' embeddings after training. This adaptive capability presents a significant advantage in dynamic real-world settings where consumer populations continually evolve, as the model maintains strong predictive performance while requiring minimal retraining.

\subsection{Ablation studies}

\subsubsection{Inclusion of categorical features.}

As several models are not explicitly designed to handle categorical features, it is important to verify that these models are not penalized by such inclusions. Results in Table \ref{tab:results-categorical} show that categorical features have overall no significant impact on their performance, with MSE varying by at most $5 \times 10^{-4}$ and MAE by at most $7 \times 10^{-4}$. Some models even show marginal improvements with categorical features (e.g., NHits exhibits lower MSE and MAE with categorical features included). This stability might suggest that the performance degradation observed in Table \ref{tab:results} is predominantly attributable to numerical features rather than categorical ones.

\begin{table}[!ht]
    \centering
    \setlength{\tabcolsep}{3pt}
    \caption{Performance ($\pm$ standard error) with and without categorical features, for models that only handle numerical time series as inputs (asterisked in Table \ref{tab:results})}
    \begin{tabular}{l c c c c c}
        \toprule
        \multirow{2}{*}{{Model}} & \multicolumn{2}{c}{{With categorical features}} & & \multicolumn{2}{c}{{Without categorical features}} \\ \cmidrule(lr){2-3} \cmidrule(lr){5-6}
        & \textbf{MSE} & \textbf{MAE} & & \textbf{MSE} & \textbf{MAE} \\ \midrule
        iTransformer  & 0.1969 {\scriptsize $\pm$ 0.0013} & \textbf{0.1966} {\scriptsize $\pm$ 0.0012} & & \textbf{0.1966} {\scriptsize $\pm$ 0.0014} & 0.1971 {\scriptsize $\pm$ 0.0012} \\
        CARD          & 0.1765 {\scriptsize $\pm$ 0.0000} & 0.1822 {\scriptsize $\pm$ 0.0001} & & \textbf{0.1764} {\scriptsize $\pm$ 0.0000} & 0.1822 {\scriptsize $\pm$ 0.0001} \\
        NHits         & \textbf{0.1763} {\scriptsize $\pm$ 0.0002} & \textbf{0.1854} {\scriptsize $\pm$ 0.0005} & & 0.1768 {\scriptsize $\pm$ 0.0002} & 0.1861 {\scriptsize $\pm$ 0.0005} \\
        PatchTST      & 0.1768 {\scriptsize $\pm$ 0.0001} & \textbf{0.1856} {\scriptsize $\pm$ 0.0001} & & \textbf{0.1767} {\scriptsize $\pm$ 0.0001} & 0.1857 {\scriptsize $\pm$ 0.0001} \\
        RLinear       & \textbf{0.1888} {\scriptsize $\pm$ 0.0000} & \textbf{0.2044} {\scriptsize $\pm$ 0.0001} & & 0.1890 {\scriptsize $\pm$ 0.0000} & 0.2048 {\scriptsize $\pm$ 0.0001} \\
        \bottomrule
    \end{tabular}
    \label{tab:results-categorical}
\end{table}

\subsubsection{Importance of different external factors.}

Table \ref{tab:football} demonstrates the significant contribution of each external factor to model performance. The experiment incorporating all external factors achieves the lowest error, while removing any category of factors leads to performance degradation. Temporal indicators emerge as the most critical component, with their removal causing the largest increase in error, followed by weather indicators and perhaps more interestingly football events. We also observe that including more external factors systematically decreases the standard error, thus making the performance less uncertain. Overall, these results quantitatively validate the hypernetwork's capacity to effectively integrate diverse external signals, capturing complex interdependencies between seemingly disparate factors and the target variable. 

\begin{table}[!ht]
    \centering
    \setlength{\tabcolsep}{3pt}
    \caption{MSE and MAE values ($\pm$ standard error) for the hypernetwork model with and without groups of external factors}
    \begin{tabular}{l c c}
        \toprule
        Removed data & \textbf{MSE} & \textbf{MAE} \\ \midrule
        Weather indicators  & 0.1768 {\scriptsize $\pm$ 0.0008} & 0.1865 {\scriptsize $\pm$ 0.0013} \\
        Date \& time indicators  & 0.1777 {\scriptsize $\pm$ 0.0010} & 0.1875 {\scriptsize $\pm$ 0.0018} \\
        Football events  & 0.1761 {\scriptsize $\pm$ 0.0004} & 0.1850 {\scriptsize $\pm$ 0.0009} \\
        $\varnothing$  & \textbf{0.1734} {\scriptsize $\pm$ 0.0002} & \textbf{0.1808} {\scriptsize $\pm$ 0.0004} \\ 
        \bottomrule
    \end{tabular}
    \label{tab:football}
\end{table}

\section{Discussion}

\paragraph{Mixed role of external factors.} The findings from our study challenge the prevailing assumption that integrating more external factors naturally enhances forecasting accuracy. Our results indicate that, for most models, the inclusion of additional external factors often leads to performance degradation. This suggests that the signal-to-noise ratio introduced by these external factors may not always be beneficial, highlighting the complexity involved in effectively leveraging such data.

\paragraph{Linearity of the forecasting process.} While the end forecast is inherently linear and may not capture complex patterns directly~\cite{dlinear}, the linear weights themselves are dynamically generated by the hypernetwork, which is nonlinear. This unique capability allows the linear model to adapt to more complex situations by tailoring weights to individual consumer behaviors, effectively making the final forecast nonlinear w.r.t. the input embeddings.

\paragraph{Adaptability.} Hypernetworks present a notable exception by using external information without compromising performance, showcasing their capability in adapting to the varying significance of different input channels. New consumer embeddings can effectively be added over time to adapt to the demand evolution, which makes this solution suitable for real-world scenario. Encoders could be used in the future to be more effective than gradient descent in order to optimize these new embeddings.

\paragraph{Future work.} Long time series embedding models~\cite{trep}\cite{timexer} could be used to create consumers embeddings optimized to serve as hypernetwork's input. This would allow even faster profile embedding without having to apply gradient descent. More complex models than simple linear models could also be considered as for the hypernetwork's output. As already suggest with MoLE models, adding simple layers to the output model could potentially increase the performance.

\section{Conclusion}

In conclusion, our investigation into leveraging hypernetworks for electrical consumption forecasting reveals their potential as a robust alternative to traditional methods. By successfully exploiting external factors without degrading model performance for a reasonable cost, hypernetworks offer a promising direction for future research, especially in applications requiring the integration of diverse data channels. The results highlight the need for continued exploration into models that effectively balance complexity and accuracy, with improvements yet to be made to optimize new consumers embeddings, overall encouraging advancements in time series forecasting, especially with real-world applications.

\section*{Acknowledgment}
The authors would like to thank Creos Luxembourg S.A. for its support and valuable feedback.

\bibliographystyle{splncs04}

\newpage

\section*{Appendix}

\begin{table}[h]
    \centering
    \setlength{\tabcolsep}{3pt}
    \caption{MSE and MAE values for different models and datasets ($\pm$ standard error). Models denoted with an asterisk * are not meant to handle categorical features: the consumer's ID embedding is provided in additional time series channels}
    \begin{tabular}{l c c c c c c c c}
        \toprule
        \multirow{2}{*}{{Model}} & \multicolumn{2}{c}{{No external factor}} & & \multicolumn{2}{c}{{Consumer ID only}} & & \multicolumn{2}{c}{{+ External factors}} \\ \cmidrule(lr){2-3} \cmidrule(lr){5-6} \cmidrule(lr){8-9}
        & \textbf{MSE} & \textbf{MAE} & & \textbf{MSE} & \textbf{MAE} & & \textbf{MSE} & \textbf{MAE} \\ \midrule
        Our method  &   -    &   -    & & 0.1771 & 0.1872 & & \underline{0.1734} & \underline{0.1805} \\
                    &        &        & & {\scriptsize $\pm$ 0.0003} & {\scriptsize $\pm$ 0.0005} & & {\scriptsize $\pm$ 0.0002} & {\scriptsize $\pm$ 0.0004} \\
                    \noalign{\medskip}
        MoLE\_DLinear & 0.1788 & 0.1899 & & 0.1798 & 0.1891 & & 0.1807 & 0.1904 \\
                    & {\scriptsize $\pm$ 0.0003} & {\scriptsize $\pm$ 0.0004} & & {\scriptsize $\pm$ 0.0000} & {\scriptsize $\pm$ 0.0001} & & {\scriptsize $\pm$ 0.0001} & {\scriptsize $\pm$ 0.0002} \\
                    \noalign{\medskip}
        MoLE\_RLinear & 0.1786 & 0.1836 & & 0.1795 & 0.1844 & & 0.1787 & 0.1839 \\
                    & {\scriptsize $\pm$ 0.0002} & {\scriptsize $\pm$ 0.0002} & & {\scriptsize $\pm$ 0.0002} & {\scriptsize $\pm$ 0.0002} & & {\scriptsize $\pm$ 0.0002} & {\scriptsize $\pm$ 0.0002} \\
                    \noalign{\medskip}
        MoLE\_RMLP    & 0.1774 & 0.1820 & & 0.1788 & 0.1832 & & 0.1778 & 0.1826 \\
                    & {\scriptsize $\pm$ 0.0001} & {\scriptsize $\pm$ 0.0003} & & {\scriptsize $\pm$ 0.0001} & {\scriptsize $\pm$ 0.0003} & & {\scriptsize $\pm$ 0.0002} & {\scriptsize $\pm$ 0.0004} \\
                    \noalign{\medskip}
        Individual RLinears*  &   -    &   -    & & 0.1741 & 0.1819 & & \textbf{0.1725} & \textbf{0.1792} \\
                    &  -  &  -  & & {\scriptsize $\pm$ 0.0021} & {\scriptsize $\pm$ 0.0041} & & {\scriptsize $\pm$ 0.0014} & {\scriptsize $\pm$ 0.0030} \\
                    \noalign{\medskip}
        RLinear*       & 0.1806 & 0.1901 & & 0.1874 & 0.1990 & & 0.1888 & 0.2044 \\
                    & {\scriptsize $\pm$ 0.0000} & {\scriptsize $\pm$ 0.0001} & & {\scriptsize $\pm$ 0.0001} & {\scriptsize $\pm$ 0.0002} & & {\scriptsize $\pm$ 0.0001} & {\scriptsize $\pm$ 0.0003} \\
                    \noalign{\medskip}
        iTransformer*  & 0.1834 & 0.1867 & & 0.1866 & 0.1894 & & 0.1969 & 0.1966 \\ 
                    & {\scriptsize $\pm$ 0.0013} & {\scriptsize $\pm$ 0.0012} & & {\scriptsize $\pm$ 0.0016} & {\scriptsize $\pm$ 0.0014} & & {\scriptsize $\pm$ 0.0016} & {\scriptsize $\pm$ 0.0018} \\
                    \noalign{\medskip}
        CARD* & 0.1759 & 0.1816 & & 0.1760 & 0.1817 & & 0.1765 & 0.1822 \\ 
                    & {\scriptsize $\pm$ 0.0000} & {\scriptsize $\pm$ 0.0001} & & {\scriptsize $\pm$ 0.0000} & {\scriptsize $\pm$ 0.0001} & & {\scriptsize $\pm$ 0.0000} & {\scriptsize $\pm$ 0.0001} \\
                    \noalign{\medskip}
        NHits* & 0.1757 & 0.1849 & & 0.1759 & 0.1851 & & 0.1763 & 0.1854 \\ 
                    & {\scriptsize $\pm$ 0.0002} & {\scriptsize $\pm$ 0.0005} & & {\scriptsize $\pm$ 0.0005} & {\scriptsize $\pm$ 0.0008} & & {\scriptsize $\pm$ 0.0004} & {\scriptsize $\pm$ 0.0006} \\
                    \noalign{\medskip}
        PatchTST* & 0.1759 & 0.1817 & & 0.1762 & 0.1822 & & 0.1768 & 0.1856 \\
                    & {\scriptsize $\pm$ 0.0001} & {\scriptsize $\pm$ 0.0001} & & {\scriptsize $\pm$ 0.0001} & {\scriptsize $\pm$ 0.0002} & & {\scriptsize $\pm$ 0.0001} & {\scriptsize $\pm$ 0.0002} \\
                    \noalign{\medskip}
        ARIMA         & 0.1780 & 0.1893 & &   -    &   -    & &   -   &  -   \\
                    & {\scriptsize $\pm$ 0.0000} & {\scriptsize $\pm$ 0.0000}  \\
        \bottomrule
    \end{tabular}
    \label{tab:results}
\end{table}

\end{document}